\documentclass[sigconf]{acmart}  

\usepackage{threeparttable}
\usepackage{multirow}
\usepackage{stfloats}  
\usepackage{booktabs}    
\usepackage{array}       
\usepackage{siunitx}
\usepackage{textcomp}     
\makeatletter
\def\@ACM@checkaffil{ 
    \if@ACM@instpresent\else
    \ClassWarningNoLine{\@classname}{No institution present for an affiliation}%
    \fi
    \if@ACM@citypresent\else
    \ClassWarningNoLine{\@classname}{No city present for an affiliation}%
    \fi
    \if@ACM@countrypresent\else
        \ClassWarningNoLine{\@classname}{No country present for an affiliation}%
    \fi
}
\makeatother  

\setcopyright{none}
\renewcommand\footnotetextcopyrightpermission[1]{}  
\usepackage{enumitem}  
\setlist[itemize]{topsep=0pt, partopsep=0pt, leftmargin=8pt} 

\begin{document}
\begin{sloppypar} 

\title{GTS-LUM: Reshaping User Behavior Modeling with LLMs in Telecommunications Industry}

\newcommand{\corrauth}{\textsuperscript{†}} 
\author{Liu Shi, Tianwu Zhou\corrauth, Wei Xu, Li Liu, Zhexin Cui, Shaoyi Liang, Haoxing Niu, Yichong Tian, Jianwei Guo} 
\affiliation{
  \institution{Huawei GTS, Xi'an, China} 
}
\email{{shiliu3, zhoutianwu1, xuwei211, liuli122, cuizhexin, liangshaoyi, niuhaoxing1, 
tianyichong, guojianwei}@huawei.com}  










\renewcommand{\shortauthors}{Shi et al.}

\begin{abstract}

As telecommunication service providers shifting their focus to analyzing user behavior for package design and marketing interventions, a critical challenge lies in developing a unified, end-to-end framework capable of modeling long-term and periodic user behavior sequences with diverse time granularities, multi-modal data inputs, and heterogeneous labels.  This paper introduces GTS-LUM, a novel user behavior model that redefines modeling paradigms in telecommunication settings. GTS-LUM adopts a (multi-modal) encoder-adapter-LLM decoder architecture, enhanced with several telecom-specific innovations. Specifically, the model incorporates an advanced timestamp processing method to handle varying time granularities. It also supports multi-modal data inputs---including structured tables and behavior co-occurrence graphs---and aligns these with semantic information extracted by a tokenizer using a Q-former structure. Additionally, GTS-LUM integrates a front-placed target-aware mechanism to highlight historical behaviors most relevant to the target. Extensive experiments on industrial dataset validate the effectiveness of this end-to-end framework and also demonstrate that GTS-LUM outperforms LLM4Rec approaches which are popular in recommendation systems, offering an effective and generalizing solution for user behavior modeling in telecommunications. 
\end{abstract}

\maketitle 


\section{Introduction} 

Over the past two decades, the telecommunications industry has witnessed significant growth, leading to a nearly saturated market with a penetration rate reaching 94\%\footnote{https://www.thebusinessresearchcompany.com/report/telecom-global-market-report}. In response to this trend, service providers have progressively shifted their focus towards the management of existing users. Despite the market saturation, retaining users remains a critical challenge---high churn rates, increased package downgrade rates, and negative net growth of newly adopted users. Thanks to vast amounts of data users generated, often reaching the Petabyte (PB) level per day, a promising approach is to analyze users' historical behavior patterns and develop predictive models to anticipate their future behavior. These insights can then be leveraged to recommend the most suitable plans or make marketing interventions, thereby improving user retention.
 
Existing approaches to User Behavior Modeling (UBM) in the telecom industry\cite{jaintelecom2021, review2024,ChurnNet2024,PPFCM2019,ALBOUKAEY2020113779, CALZADAINFANTE2020113553} primarily follow traditional machine learning techniques, such as logistic regression, XGBoost and DNN. These methods, while effective in certain contexts, are heavily dependent on customized feature engineering processes and exhibit limitations in handling complex data structures and generalizing across diverse scenarios. Recent advancements in recommendation systems (RS) on Internet platforms \cite{HLLM2024, LLMKGRec2024, NoteLLM2024, NoteLLM22024, HSTU2024, LEARN2024} also offer valuable insights for the field of UBM. These systems harness Large Language Models (LLMs) to understand users' latent interests and retrieve the most likely purchased items, forming the new LLM4Rec paradigm. LLM4Rec can be broadly categorized into two approaches\cite{LEARN2024}: (1) LLM-to-Rec, which utilizes LLMs to generate high-quality embeddings from users' behavior sequences to enhance recommendation models\cite{HLLM2024,NoteLLM2024,LEARN2024}; (2) Rec-to-LLM, which transforms recommendation tasks into a dialogue format, using LLMs to answer users' next-tiem behavior\cite{Text2Tracks2024, ECR2024}. Due to LLM's superior capabilities in semantic comprehension, generation and generlization across different tasks with limited labeled data, LLM4Rec has achieved state-of-the-art (SOTA) performance on open datasets and demonstrated significant enhancements in real-world experiments.  

\begin{figure}[t]
  \centering
  \includegraphics[width=1\linewidth]{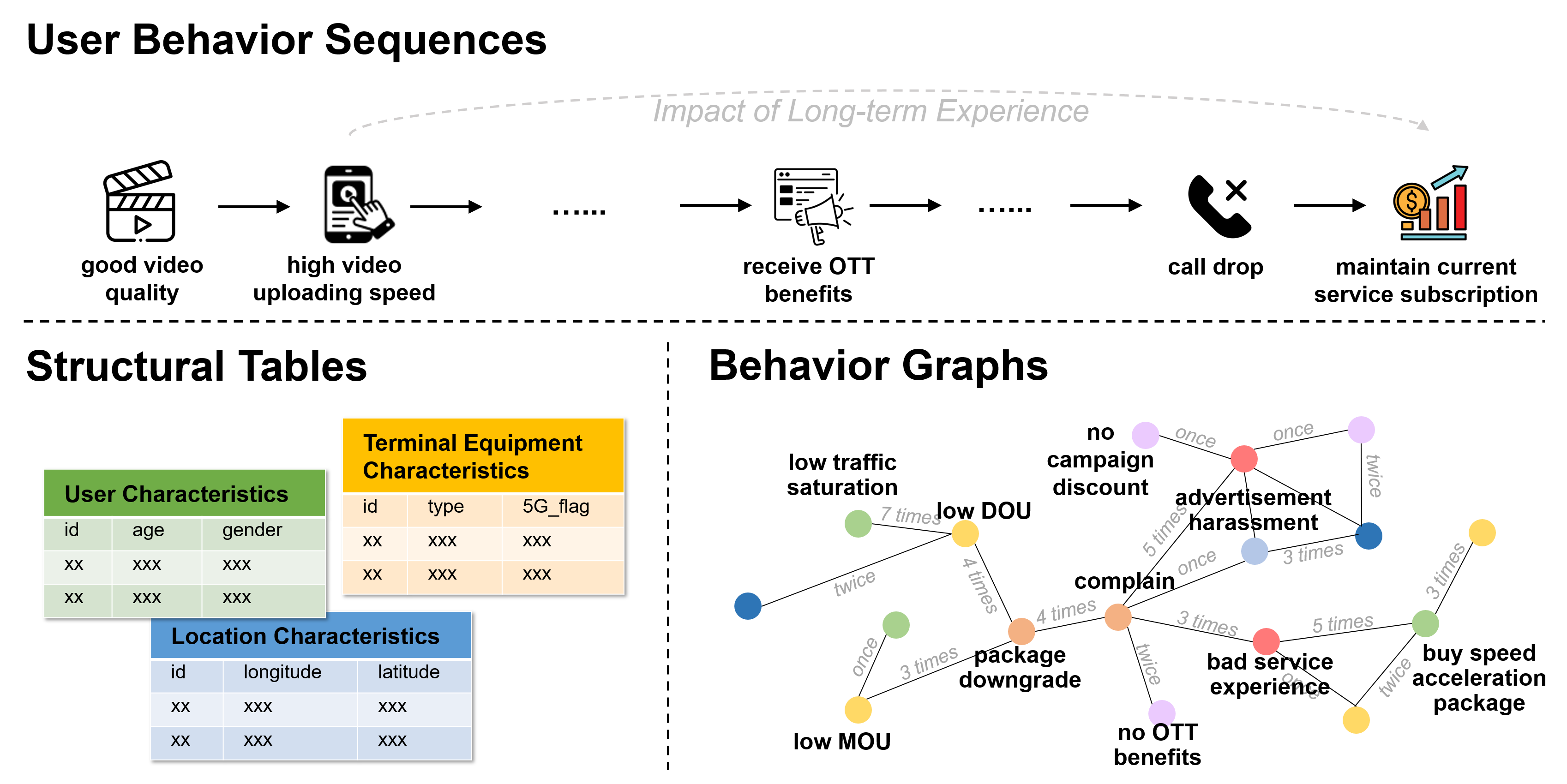}
  \caption{Diverse Formats of User Behavior Data in the Telecommunications Industry}   
  \label{fig_user_behavior}
\end{figure}
Despite the success of LLM4Rec paradigm, their application to telecommunications settings faces significant challenges. The first distinction is users' \textbf{decision-making mechanism}: telecom users are influenced more by \textit{long-term experience} than recent interests and have \textit{periodic preferences}, as services are deeply integrated in their daily routines over extended periods. As shown in Figure \ref{fig_user_behavior}, a user who has consistently experienced reliable network services and reasonable monthly pricing plans is unlikely to switch to a competing service provider due to a recent call drop. In LLM4Rec, systems typically employ a recency-based sampling strategy\cite{LEARN2024} because LLMs often struggle with processing long input contexts, leading to high inference computational costs. These approaches, which, while effective in other domains, may hurt prediction accuracy in telecom scenarios where long-term context is crucial. 

The second distinction lies in the \textbf{data characteristics} inherent in the telecommunications industry. The information of user behaviors can be found in \textit{diverse formats}, such as structural tables and beahvior co-appearance graphs, rather than merely simple item purchase histories in RS (see Figure \ref{fig_user_behavior}). This complexity necessitates a unified predictive framework capable of integrating heterogeneous data inputs while aligning semantic information across modalities, which is not common in previous RS literature. Furthermore, targets also have significant \textit{heterogeneity}, including churn, package upgrades/downgrades, marketing campaign acceptance, etc. While the underlying drivers of these targets may vary dynamically, existing LLM4Rec methods often fail to deeply capture the interactions between these targets and users' historical behaviors. Additionally, user behaviors in telecom exhibit \textit{different time granularities}; for instance, bill payments are recorded monthly, while location-based service requests occur at a per-second level. Since the sequence input is critical to performance, it is essentical to design a delicate timestamp encoding mechanism that accounts for periodic and sequential relationship---an aspect often overlooked in LLM4Rec approaches.

To address these challenges, this paper proposes a novel \textbf{L}arge \textbf{U}ser behavior \textbf{M}odeling method tailored for \textbf{T}elecommunications industries, termed GTS-LUM. The approach begins with a refined timestamp process for user sequences. Each day is divided into several fixed-length intervals, with each interval represented by a semantic text descriptor (e.g., ``Wednesday morning rush hour'') and a fine-grained time mark. The semantic text captures coarse-grained periodic patterns, while the time mark facilitates the modeling of relative and causal relationships between behaviors across different time segments. Next, we introduce a target-aware sequence modeling framework\cite{DIN2018, DIEN2018}, to address target heterogeneity. Unlike the interleaving of items and actions in HSTU\cite{HSTU2024}, our approach positions the target at the beginning of user behavior sequences, enabling deeper interaction between historical behaviors and the target. 
 
Our model architecture also employs spectral clustering to generate semantic tokens for each behavior in the sequence and utilizes multi-modal encoders to transform tabular and graph data into embeddings. To bridge the gap between generated semantic IDs, which carry open-world meanings, and multi-modal representations, which encapsulate rich business insights, we leverage a Q-former\cite{Blip22023,ILM2024} to make an alignment and construct a fused embedding. Notably, we introduce additional training tasks for Q-former, which includes sequence-text matching, sequence-text contrastive learning, and sequence-grounded text generation, enabling it to function as both an alignment mechanism and a compressor to effectively capture users' long-term interests. The fused embeddings, combined with timestamp tokens, are processed by an LLM decoder to produce user embeddings, which can be adapted to downstream tasks via task-specific heads.

To evaluate the effectiveness of GTS-LUM, we conducted experiments on an industrial dataset. The results demonstrate that GTS-LUM surpasses previous state-of-the-art (SOTA) methods in performance, underscoring its practical applicability.

To summarize, the major contributions of this paper are as follows:
\begin{itemize}  
  \item We redefine the user behavior sequence modeling process in the telecommunications industry by introducing a dedicate timestamp process and a target-aware modeling framework to enhance temporal and contextual understanding.
  \item We propose a novel modeling architecture that integrates semantic tokenization methods and multi-modal encoding methods to unify inputs from diverse data formats and employs Q-former to serve a dual purpose: an alignment to bridge the gap between semantic information and business-specific insights, and a compressor to capture users' long-term  interests.
  \item To the best of our knowledge, GTS-LUM is the first to adopt the LLM4Rec paradigm for end-to-end tasks in the telecommunications industry. Experiments on real-world datasets validate its practical applicability and effectiveness.
\end{itemize}

\section{Related Work} 

\subsection{Predictive Methods in Telecommunication Industry}

Previous literature has highlights significant advancements in predicting user churn within the telecommunications industry\cite{jaintelecom2021, review2024}. Early studies primarily employed rule-based methodologies, which have gradually transitioned to machine-learning approaches, such as logistic regression, support vector machines, and tree-based techniques\cite{jaintelecom2021}. A critical challenge in this domain is the selection of the most relevant features that contribute to predicting user churn. Recently, deep learning methods have been adopted to model user behavior in telecommunications settings, for example, \cite{PPFCM2019} and ChurnNet\cite{ChurnNet2024}. While these models have demonstrated improvements in accuracy and complexity, they often lack flexibility when dealing with heterogeneous targets and multi-modal inputs.

In contrast, our work pioneers the use of Large Language Model (LLM)-based methods for user behavior modeling. This approach not only enhances predictive performance but also provides greater flexibility to accommodate diverse downstream tasks, addressing key limitations of prior research.
 
\subsection{User Behavior Modeling in LLM4Rec} 

Large Language Models (LLMs) have garnered significant attention for their advanced representation and reasoning capabilities, which have been increasingly leveraged for user behavior modeling. Specifically, LLMs can generate high-quality embeddings that capture users' latent interests, thereby providing meaningful insights for subsequent recommendation tasks\cite{HLLM2024,NoteLLM2024,LEARN2024}. Additionally, LLMs are capable of interpreting natural language descriptions of user behavior interactions and predicting future actions\cite{Text2Tracks2024, ECR2024}.  

As recommendation systems represent the primary application scenario for user behavior modeling, the quality of input becomes a critical determinant of recommendation success. Recent advancements integrate multimodal data---such as text, images, and videos---to enhance performance.\cite{NoteLLM22024, ILM2024, RecGPT4V2024, MLLMMSR2025}. For instance, NoteLLM2\cite{NoteLLM22024} pioneers end-to-end fine-tuning with text and image modalities, while ILM\cite{ILM2024} combines collaborative filtering signals and text descriptions, utilizing a Q-former for text-aligned item representations inspired by Blip-2\cite{Blip22023}. These models typically consist of an encoder (e.g., a vision model or a collaborative filtering model), a connector, and an LLM-decoder, ensuring training and fine-tuning efficiency. Our approach builds upon the ILM framework while extending its capabilities to accommodate a broader range of data modalities, including tables and co-appearance graphs. Additionally, we introduce a behavior tokenization technique to more effectively capture semantic similarities among diverse behaviors. This enhancement significantly improves the quality of user embeddings.
 
\subsection{Item Tokenization in Generative Retrieval}

Inspired by generative retrieval, there is a novel paradigm known as generative recommendation which tokenizes each item and reformulates the recommendation task as a token sequence generation problem. The paradigm is gaining popularity because it leverages the rich content information in item text descriptions and avoids possible challenges of sparse and imbalanced training data\cite{STORE2024},  different from traditional methods that assign an atomic identifier to each item. Two primary techniques exist for item tokenization. The first technique, pioneered by TIGER\cite{Tiger2023} uses residual quantization to generate low-dimensional codeword tuples as semantic identifiers. This approach effectively captures the information of each item in a hierarchical manner, ranging from coarse to fine-grained representations. The second technique is hierachical clustering \cite{RecForest2022,HowtoIndexItem2023,Colarec2024,EAGER2024}, which groups similar items iteratively into clusters, with the root-to-leaf path defining the item's semantic ID. A key challenge in both methods is bridging the gap between general-domain corpora and item-specific content within the recommendation scenario. To address this, researchers have replaced the input with collaborative signals, such as user-item interaction graphs\cite{HowtoIndexItem2023,Colarec2024,MMGRec2024,LETTER2024}. However, few studies have integrated both collaborative information and item text description simultaneously, except those in \cite{EAGER2024, LETTER2024, li2024semanticconvergenceharmonizingrecommender}. In our approach, we also consider these two streams of information as inputs but treat them as multi-modal data. Unlike previous methods, we leverage a Q-former\cite{Blip22023,ILM2024} to learn more effective and unified representations for each token, capturing the complementary strengths of both collaborative and content-based signals. 

\section{Method}
In this section, we first introduce the problem formulation, and then present the architecture of Large User Behavior Modeling method for Telecommunications industries (GTS-LUM). Finally, we introduce our training strategy.

\subsection{Problem Formulation}

We investigate a sequential prediction problem, analogous to sequential recommendation in the literature. For a user $u\in\mathcal{U}$, his/her historical behavior is represented in chronological order as $U=\{b_1,b_2,..., b_n\}$, where $n$ denotes the sequence length and $b_j\in\mathcal{B}\,(1\leq j\leq n)$ represents the $j$-th behavior. The corresponding timestamps are captured in $T=\{t_1, t_2, ..., t_n\}$, with each $t_j$ expressed details including year, month, day, hour, minute, and second. Our primary objective is to utilize the behavior history $U$ and corresponding time sequence $T$ to predict the next behavior $b_{n+1}$. 

\subsection{Model Architecture}

\begin{figure*}[t]
  \centering
  \includegraphics[width=1\linewidth]{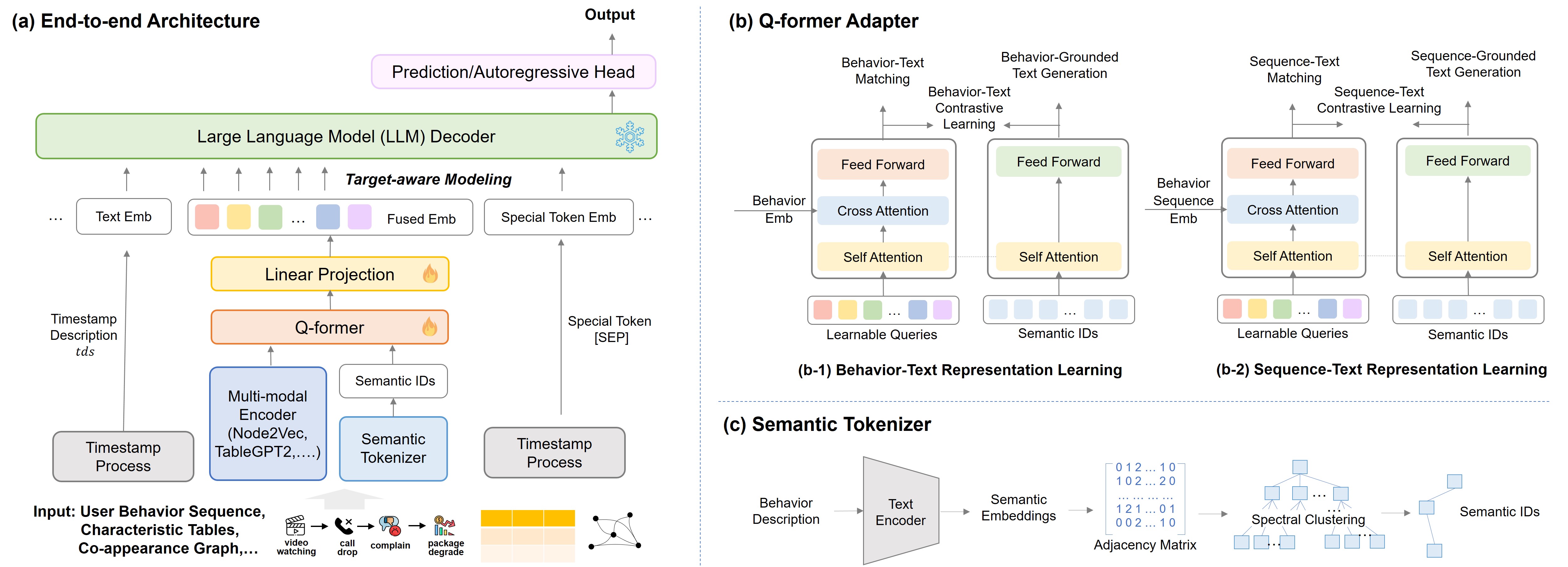}
  \caption{Overview of GTS-LUM}   
  \label{figure_overview}  
\end{figure*}

As shown in Figure \ref{figure_overview}(a), GTS-LUM allows multi-modal data as inputs, including behavior sequences, characteristic tables, and co-appearance graphs. We initially develop a telecom-specific timestamp process for $T$. A semantic tokenizer converts textual behavior into semantic IDs, while a pretrained multi-modal encoder transforms table and graph data into embeddings. We align these tokens and embeddings using a Q-former and feed them into a linear projection layer. Finally, we model user profile with a target-aware LLM decoder. The output embedding can be connected with either an autoregressive head or a prediction head to fit into different downstream tasks.

\subsubsection*{\textbf{Timestamp Process}} 
In this paper, we aim to develop a timestamp process method with both coarse-grained and fine-grained information to model the periodic and relative relationships among behaviors.

Following \cite{HSTU2024}, we first select the longest time series as the main sequence and partition each day into same-range intervals to overcome diverse temporal granularitiers inherent in user behavior. Behaviors occurring within the same interval are assigned a shared time ID, meaning behavioral co-occurrences. Beyond the time ID, each interval is enriched with semantic context using a combination of two dimensions: (1) weekday vs. weekend, and (2) time categories such as morning rush hour, normal morning hour, noon, evening rush hour, normal evening hour, and late night. This design is motivated by the observation that user preferences exhibit periodic variations \cite{Zhang_2023}, enabling Large Language Models to better capture dependencies in user behavior. For example, the interval from 9:00 to 9:30 on a Monday morning, classified as a weekday morning rush hour, is typically associated with high network usage due to commuting and work-related activities. In contrast, the same interval on a Sunday morning, labeled as a weekend normal hour, may reflect entirely different behavioral patterns. 

When reshaping the behavior sequence using semantic descriptions and IDs, we prepend the semantic description $tds_{k}$ to co-occurred behaviors during interval $k$. Additionally, we append a special token [SEP] to separate intervals with different IDs. Consequently, the input sequence is reformulated as: 
\begin{equation}
  U=\{tds_1, b_1, b_2,...,b_j, \text{[SEP]}, tds_2, b_{j+1}, ..., \text{[SEP]}, ..., b_n\}.
\end{equation}

\subsubsection*{\textbf{Multi-modal Alignment}}  
 
Given the behavior $b_{j}$ in the sequence $U$, there is a textual description with semantically meaningful information. Departing from conventional methods that assign a discrete atomic identifier to each behavior \cite{tang2018personalizedtopnsequentialrecommendation,P52022,ijcai2019p600}, we propose that semantically analogous behaviors should be represented by structurally similar IDs\cite{EAGER2024,Colarec2024,RecForest2022,HowtoIndexItem2023}. To achieve this, we first leverage a general-purpose text encoder (e.g., BGE-M3\cite{bgem32024}) to convert the textual descriptions of behaviors into high-dimensional vector representations. These embeddings are then convert to a adjacency matrix, followed by a standarad spectral clustering algorithm. The clustering result can be formulated as a spectral tree, where the node sequence of the path from the root to a leaf is defined as the semantic ID of the corresponding behavior, denoted as $b_{j}^\mathbf{S}$. Figure \ref{figure_overview}(c) indicates the semantic ID generation process.  

While above tokens can capture behavior semantics, telecom user modeling necessitates complementary business insights derived from multi-modal behavior data. These include signals such as behavior co-appearance graphs, location and terminal equipment characteristics tables.  Incorporating these information into user embeddings has the potential to significantly enhance the quality of user representations.

However, a fundamental gap exists between general-domain semantic information and domain-specific business signals. To bridge this gap, we propose a Q-former-based architecture\cite{Blip22023,ILM2024} which aligns business and semantic information through multi-modal attention mechanisms. As illustrated in Figure \ref{figure_overview}(b), we first derive business embeddings (i.e., $b_{j}^\mathbf{E}$) via domain-adaptive encoders: collaborative filtering signals, such as behavior co-appearance patterns, are encoded using Node2Vec \cite{node2vec2016}, while structured attributes (e.g., service plans, device types) are processed based on TableGPT2\cite{tablegpt22024}. These multi-modal embeddings, along with semantic tokens, are then fed into the Q-former module. Within the Q-former, cross-attention layers enable learnable queries to dynamically extract business features that are most relevant to the semantic context, while shared self-attention layers establish latent alignments between business attributes and semantic ID components. The resulting query representations are passed through a feed-forward layer to produce fused behavior embeddings $b_j^\mathbf{F}$. Note that although EAGER\cite{EAGER2024} also utilizes the two-streams of information, their parallel decoding approach struggles to model deep, fine-grained interactions between semantic and business signals.  
 
\subsubsection*{\textbf{Target-aware Modeling}}

In telecommunications industries, where predictive labels are heterogeneous such as churn and package upgrades/downgrades, it is crucial to highlight the most relevant historical behaviors associated with the target label. Drawing inspiration from prior work \cite{DIN2018,HSTU2024}, we adopt a target-aware modeling approach, enabling a deep integration between user interests and candidate items to form a comprehensive user representation. However, our approach introduces a key structural difference: instead of appending the target behavior $b^\mathbf{F}_{\text{tgt}}$ to the end of the historical behavior sequence, we position it at the beginning. This reordered sequence is then fed into a LLM decoder to generate high-order cross-features, with the representation of the last token extracted as the user embedding.

This reordering, though seemingly minor, leverages the causal attention mechanism of LLM decoders to fundamentally alter the interaction dynamics between the target and historical behaviors. As formalized in Eq.(\ref{eq_w_tgt_aw}), placing $b^\mathbf{F}_{\text{tgt}}$ at the start enables each subsequent behavior $j$ to compute attention over all preceding elements, including the target:
\begin{equation}
  y_j = \sum_{i={\text{tgt}, 1,..,j}} \text{attn}(i,j)v_i \quad  \text{where}\,j\in\{\text{tgt}, 1,..,n\},\label{eq_w_tgt_aw}
\end{equation}
where $y_j$ denotes the $j$-th behavior's representation after one casual attention layer and $v_i$ denotes the learnable vector with information of behavior $i$. In contrast, existing methods \cite{HSTU2024,HLLM2024} that append $b^\mathbf{F}_{\text{tgt}}$ (i.e., Eq. \ref{eq_wo_tgt_aw})  compute attention for historical behaviors independently of the target, only integrating them at the final step:
\begin{equation}
  y_j=\begin{cases}
    \sum_{i={1,..,j}} \quad\,\text{attn}(i,j)v_i\quad  \text{where}\,j\in\{1,..,n\}\\
    \sum_{i={1,..,\text{tgt}}}\, \text{attn}(i,\text{tgt})v_i \quad  \text{where}\,j=\text{tgt}
  \end{cases} \label{eq_wo_tgt_aw}
\end{equation} 
Therefore, target-aware modeling in our paper allows iterative refinement of user embeddings, as each historical behavior’s influence is progressively modulated by the target. Furthermore, it can better capture users' long-term experience, which is essential in telecommunication industry.  

\subsection{Model Training and Fine-tuning}

In this section, we demonstrate the two phases of our purposed GTS-LUM model training strategy.

\subsubsection*{\textbf{Phase \uppercase\expandafter{\romannumeral1}: Q-former Pre-training}}

In Phase 1, we pretrain the Q-former adapter to generate representations that align business and semantic information. Following \cite{Blip22023,ILM2024}, our training process incorporates three alignment tasks: behavior-text contrastive learning task, behavior-text generation task, and behavior-text matching task, with corresponding loss functions, see Figure \ref{figure_overview}(b-1). 

Additionally, we introduce three novel tasks, including sequence-text matching task, sequence-text contrastive learning task and sequence-text generation task, shown in Figure \ref{figure_overview}(b-2). Inspired by MLLM studies, where an image is similar to a windowed slice of user behavior sequences, we can treat each behavior as a patch and design an alignment task between the sequence and text. This approach is particularly reasonable in telecom settings, where a behavior sequence, such as a user streaming a live sports event during peak hours, may exhibit meaningful patterns. For instance, the user experiences video freezing, resolution drops, and eventual playback failure, leading to session abandonment. This sequential degradation---buffering, resolution drops, and failure---can be represented as a ``poor video quality'' scenario caused by network instability. The sequence-text conotrastive learning task, for instance, distincts mainly from the original one in \cite{ILM2024} is that the output representation of query tower $\mathcal{F}_q$ is based on the vector of $K$ behavior embeddings $[b^{\mathbf{E}}_1, b^{\mathbf{E}}_2, ... b^{\mathbf{E}}_K]$, that is,
\begin{equation}
  [h_1,h_2, ..., h_Q] = \mathcal{F}_q([q_1, q_2, ..., q_Q], [b^{\mathbf{E}}_1, b^{\mathbf{E}}_2, ... b^{\mathbf{E}}_K]).
\end{equation}
After compute the text tower representation, we calculate the loss similar to the original task.

\subsubsection*{\textbf{Phase \uppercase\expandafter{\romannumeral2}: Model Training}}

In Phase 2, we employ self-supervised contrastive learning for end-to-end training. Specifically, the embedding of the $j$+1-th behavior serves as the positive sample for the sequence up to timestep $j$. Due to target-aware modeling, we reorder the sequence by positioning the positive sample at the forefront and input it into the LLM decoder to generate the corresponding positive embedding. Negative samples are derived from target behavior embeddings of other users within the same batch, yielding $M$ negatives, which are similarly processed to obtain negative embeddings.  Then, we compute the contrastive loss following \cite{HLLM2024} which uses InfoNCE\cite{infonce2018} as a criterion.  

During training, we optimize the parameters of the Q-former and the linear projection layer while keeping the parameters of the LLM decoder frozen.

\section{Experiment} 

In this section, we begin by assessing the performance of GTS-LUM on the industry dataset to prove its feasibility. Although the model training process consists of two phases, our experiments focus exclusively on the performance of Phase II, specifically evaluating the effectiveness of GTS-LUM's architecture when built upon a pre-trained Q-former. Subsequently, we conduct an ablation study to validate the necessity of the modules integrated into GTS-LUM.  

\subsection{Experiment Setup}
 
\subsubsection*{\textbf{Dataset}} We evaluate GTS-LUM on an industrial dataset which captures users' daily behavior records, including call logs, data usage and service interactions. The detailed statistics of the training and testing dataset are shown in Table \ref{dataset}. By integrating a autoregressive head with the user representations generated by the LLM decoder, we target OTT-service recommendation as the downstream task, a critical business scenario for customer retention and revenue optimization.

\begin{table}  
  \centering 
  \begin{tabular}{cccc}
    \toprule
    Dataset & users & avg. seq-length & time span\\
    \midrule 
    Train & 193799 & 4832 & 3 months \\ 
    Test & 18973 & 4765 & 3 months \\
    \bottomrule
  \end{tabular}  
  \caption{Details of Industry Datasets}
  \label{dataset}
\end{table}

\subsubsection*{\textbf{Baselines}} For comparison purposes, we select several state-of-the-art models as baselines:

\textbf{HSTU}\cite{HSTU2024}: This model introduces a unified framework that encodes features into time series, thereby reformulating recommendation problems as sequential transduction tasks. In our experiment, we regard HSTU as our benchmark model. To have a fair comparison, we reproduce HSTU to utilize the same hidden size and number of layers as TinyLlama-1.1B, following HLLM\cite{HLLM2024}.

\textbf{HLLM}\cite{HLLM2024}: This model proposes a novel hierarchical architecture that connects two large language models to capture item embeddings and user representations, respectively. We consider two scales of HLLM in our experiment: HLLM-0.5B and HLLM-1B. Specifically, we use Qwen2.5-0.B for HLLM-0.5B and TinyLlama-1.1B for HLLM-1B, and each model is utilized for both item and user representations within the architecture.

\subsubsection*{\textbf{Evaluation Metrics}}

The performance is assessed based on two standard metrics: top-$K$ Normalized Discounted Cumulative Gain (NDCG@$K$) and top-$K$ Recall (R@$K$), with $K$ set to 5, 10, 50 and 200, ensuring a fair comparison with baseline models. Following the standard evaluation protocol \cite{kang2018selfattentivesequentialrecommendation}, we employ a leave-one-out strategy to partition each datasets into training and testing sets. Specifically, the last item in each sequence is used for testing and the remaining items for training.

\subsubsection*{\textbf{Implementation Details}}

In GTS-LUM, the timestamp processing involves partitioning time intervals into fixed regions of 15 minutes. We employ BGE-M3 \cite{bgem32024} as the text encoder during the process of semantic tokenization. Additionally, we use Node2Vec \cite{node2vec2016} and TableGPT2\cite{tablegpt22024} as multi-modal encoder to generate embeddings from behavior co-occurrence graph and characteristic tables in our industry dataset, respectively. The LLM decoder is implemented using TinyLlama-1.1B\cite{tinyllama2024} for a fair comparison, which provides robust generative capabilities for sequence modeling and prediction tasks. 

Our model is trained on 200 $\times$ 64GB Ascend 910B2 NPUs for 5 epochs. During the training process, the batch size per NPU is set to 8. User behavior sequences are sampled according to their importance ratio, with each sequence truncated to a maximum length of 240 tokens.  The learning rate is fixed at 
1e-4, and the contrastive learning framework employs a positive-to-negative sample ratio of 1:512.

\subsection{Experiment Results}
 
\begin{table*}[t]  
  \centering 
  \begin{tabular}{
    >{\centering\arraybackslash}p{2cm}  
    *{8}{>{\centering\arraybackslash}p{1.3cm}}   
    >{\centering\arraybackslash}p{2cm}   
    }
    \toprule
    Methods & R@5 & R@10 & R@50 & R@200 & NDCG@5 & NDCG@10 & NDCG@50 & NDCG@200 & Impv.(avg)\\
    \midrule 
    HSTU-1B & 2.09 & 2.97 & 7.18 & 7.94 & 1.77 & 2.06 & 3.08 & 3.27 & 0.00$\%$\\
    HLLM-0.5B & 2.75 & 3.82 & 9.36 & 9.42 & 2.30 & 2.64 & 3.86 & 3.86 & 26.49$\%$\\
    HLLM-1B & 3.89 & 5.38 & 9.14 & 9.19 & 3.22 & 3.77 & 4.60 & 4.61 & 57.93$\%$\\
    Ours & \textbf{4.94} & \textbf{6.15} & \textbf{10.42} & \textbf{10.99} & \textbf{4.56} & \textbf{4.95} & \textbf{6.71} & \textbf{7.14} & \textbf{107.86$\%$}\\
    \bottomrule
  \end{tabular}  
  \caption{Experimental Results ($\%$)}
  \label{table_results}
\end{table*}

As evidenced by the experimental results in Table \ref{table_results}, the proposed GTS-LUM framework demonstrates consistent superiority over baseline models across all evaluated performance metrics. It's clear that although typical recommendation models achieve competitive results on open benchmarks, they exhibit significant performance degradation when applied to industrial-scale datasets characterized by domain-specific complexity. Specifically, our results show that GTS-LUM achieves statistically significant improvements, outperforming HSTU by an average of 107.86\% and surpassing HLLM-1B by 31.38\%. These improvements highlight the importance of capturing multi-modal inputs and the necessity of aligning open-world knowledge with specific business knowledge.

\subsection{Ablation Study}

In this section, we conduct ablation studies to assess the impact of GTS-LUM's key components, including timestamp processing, the Q-former architecture, and target-aware modeling, to demonstrate their necessity.

\subsubsection*{\textbf{Impact of Timestamp Processing}}

To evaluate the impact of timestamp processing in GTS-LUM, we compare results with a variant that excludes the timestamp processor (``w/o timestamp'') and three alternative methods commonly used in prior literature:
\begin{enumerate}
  \item[1. ] ``w txt emb'', which integrates textual descriptions of each timestamp into the behavior sequence, forming $U=\{t_1, b_1, t_2, b_2, ..., t_n, b_n\}$.
  \item[2. ] ``w comp emb'', following HLLM\cite{HLLM2024}, which decomposes timestamps into six components (year, month, day, hour, minute, second), transforms them into embeddings,  The embeddings are concatenated and fed into a Multi-Layer Perceptron (MLP) to generate a timestamp embedding sequence $T=\{t^\mathbf{E}_1, t^\mathbf{E}_2, ..., t^\mathbf{E}_n\} $, matching the length of the user behavior sequence. 
  \item[3. ] ``w pos emb'', which also groups behaviors within the same intervals and assigns the same position ID to each group. It then generates a position embedding sequence $P=\{p^\mathbf{E}_1, ,..., p^\mathbf{E}_1, p^\mathbf{E}_2, ..., p^\mathbf{E}_2, ..., p^\mathbf{E}_n, ..., p^\mathbf{E}_n\}$ based on the model's inherent relative position encoding approach.
\end{enumerate}

\begin{table}[t]
  \begin{tabular}{lcccc}
    \toprule
    Methods & R@5 & R@10 & NDCG@5 & NDCG@10 \\
    \midrule
    w/o timestamp & 2.02 & 3.28 & 1.56 & 1.96 \\
    w txt emb & 3.36 & 4.77 & 3.18 & 3.49 \\
    w comp emb (HLLM) & 3.47 & 4.55 & 2.80 & 3.04 \\
    w pos emb & 4.09 & 5.16 & 3.96 & 4.16 \\
    Ours & \textbf{4.94} & \textbf{6.15} & \textbf{4.56} & \textbf{4.95} \\
    \bottomrule
  \end{tabular} 
  \caption{Results of Ablation Study: Improvement of Timestamp Processing Methods}
  \label{tbl_ablation_timestamp}
\end{table}

The comparative analysis, as detailed in Table \ref{tbl_ablation_timestamp}, reveals that incorporating timestamp processing significantly enhances performance, with the average improvement of 144.13\%. This is because our approach leverages text descriptions to capture periodic behavior patterns, with a special token directing the model's focus to behaviors within the same interval. Notably, while integrating text descriptions alone or using component embeddings yields marginal improvements, the positional embedding method closely matches our timestamp process. This is attributed to the positional embedding's capacity to introduce additional learnable vectors, thereby enriching the model's representational power.

\subsubsection*{\textbf{Impact of Multi-modal Alignment}}

\begin{figure}[t]
  \centering
  \includegraphics[width=1\linewidth]{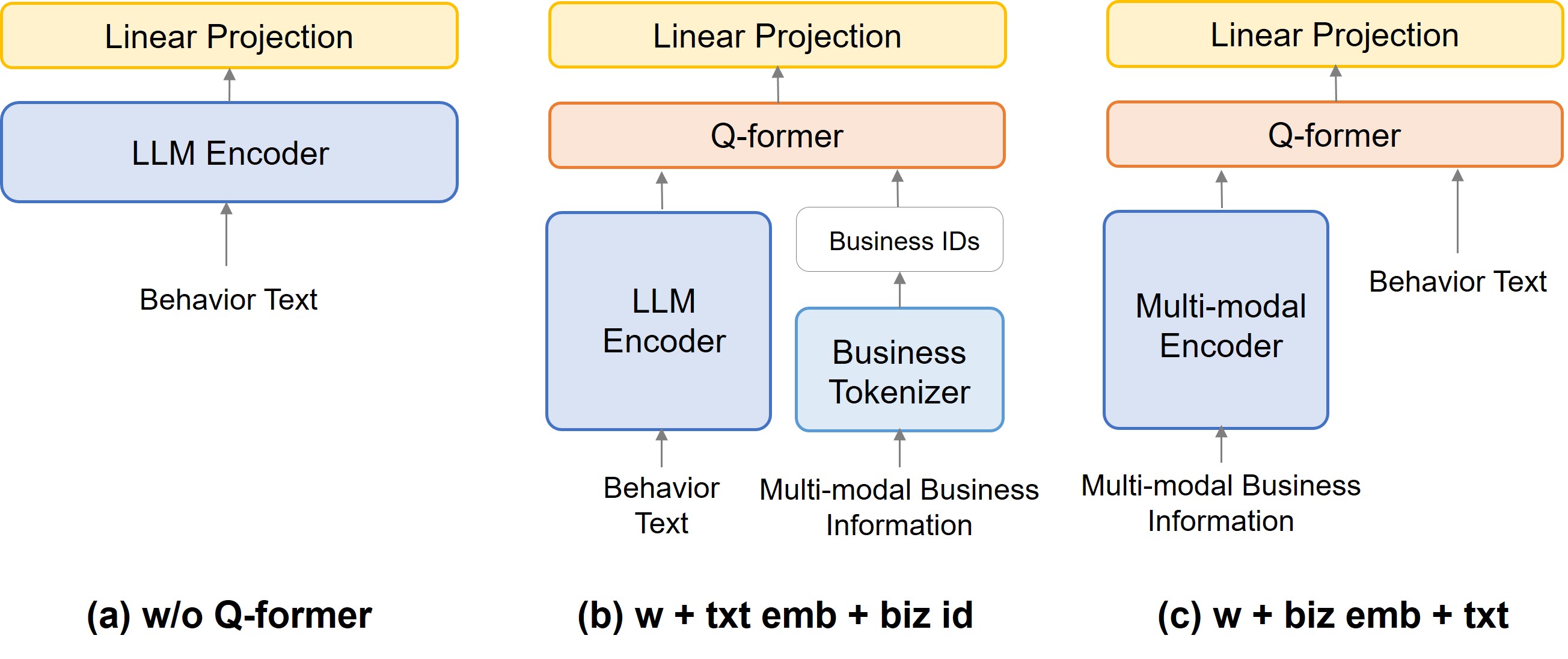}
  \caption{Depiction of Variants in the Ablation Study of Multi-modal Alignment}   
  \label{figure_ablation}  
\end{figure}

To investigate the impact of the Q-former structure and the use of multi-modal data in GTS-LUM, we conduct comparative experiments. Specifically, we compare GTS-LUM with a variant that excludes the Q-former and uses an LLM encoder to process the textual description of each behavior in the sequence (``w/o Q-former''). Figure \ref{figure_ablation}(a) provides a clear illustration of this variant. 

Additionally, we examine two alternative approaches that incorporate both multi-modal business information and behavior text as input but align them differently. Figure \ref{figure_ablation}(b) represents the approach ``w txt emb + biz id'', where we use an LLM encoder to embed the behavior textual content and a business tokenizer to encode the multi-modal business information. The behavior text embeddings interact with learnable queries through a cross-attention mechanism and align with business IDs through a co-used self-attention mechanism. Note that the semantic tokenizer introduced in GTS-LUM, with the spirit of spectral clustering, is also suitable for multi-modal inputs such as behavior-cooccurrence graphs. Figure \ref{figure_ablation}(c) depicts the approach ``w biz emb + txt''. Unlike GTS-LUM, this method directly uses the behavior textual token as Q-former input without extracting semantic information from the textual content. 

\begin{table}[t]
  \begin{tabular}{lcccc}
    \toprule
    Methods & R@5 & R@10 & NDCG@5 & NDCG@10 \\
    \midrule
    w/o Q-former & 4.34 & 6.03 & 3.59 & 4.13 \\
    w txt emb + biz id & 4.09 & 5.75 & 3.47 & 4.00 \\ 
    w biz emb + txt & 4.73 & 6.10 & 4.36 & 4.78 \\
    Ours  & \textbf{4.94} & \textbf{6.15} & \textbf{4.56} & \textbf{4.95} \\
    \bottomrule
  \end{tabular} 
  \caption{Results of Ablation Study: Improvement of Q-former Architecture}
  \label{tbl_ablation_Q_former}
\end{table}

The results presented in Table \ref{tbl_ablation_Q_former} illustrate the outstanding performance achieved by integrating multi-modal information compared to models relying solely on textual behavior data. Besides, by comparing different structures of composing multi-modal information (i.e., see Figure \ref{figure_ablation}(b)), our analysis further reveals that facilitating interaction between multi-modal embeddings and learnable queries via a cross-attention mechanism is more effective. This approach is particularly beneficial as it aligns with the core principle of the Q-former, which is to distill the most pertinent information from complex data sources to enhance representational quality. In industrial applications, while business-related information holds significance, the semantic content embedded within textual descriptions is crucial for understanding the interrelationships among various behaviors.

Moreover, when comparing the first line in Table \ref{tbl_ablation_Q_former} with the result of HLLM-1B in Table \ref{table_results}, we can find that introducing a lightweight projector layer between two LLMs significantly enhances cross-model compatibility. Unlike HLLM's hierachical architecture, which constrains both Item LLM and User LLM to identical hidden dimensions, our projector may perform dimensionality reduction through projection operations, thereby reducing computational complexity but keeping flexibility while improving the model's ability to capture user requirements.

\subsubsection*{\textbf{Impact of Target-aware Modeling}}

In Section 3.2, we proposed a novel target-aware modeling method that places the target behavior at the front of the behavior sequence. In this section, we conduct another experiment on a small dataset due to the resource constraint, where sampling 20K users from the original industrial dataset, to evaluate the impact of this placement on the final results. Table \ref{tbl_ablation_target_aware} summarizes the findings. Specifically, ``w/o target-aware'' indicates that the user representation vector is generated without incorporating target behavior information, while ``w target at end'' refers to the case where the target behavior is placed at the end of the sequence before inputting it to the LLM decoder.

\begin{table}[t]
  \begin{tabular}{lcccc}
    \toprule
    Methods & R@5 & R@10 & R@50 & R@200 \\
    \midrule
    w/o target-aware & 3.13 & 3.54 & 6.44 & 6.49 \\
    w target at end & 3.43 & 4.19 & 6.09 & 6.60 \\ 
    Ours & \textbf{3.78} & \textbf{4.20} & \textbf{6.53} & \textbf{6.99} \\ 
    \bottomrule
  \end{tabular} 
  \caption{Results of Ablation Study: Target-aware Modeling}
  \label{tbl_ablation_target_aware}
\end{table}

Our findings highlight the critical role of incorporating target behavior in user representation learning, aligning with prior studies \cite{DIN2018,DIEN2018}. Unlike existing approaches that require architectural modifications, we demonstrate that simply positioning the target behavior at the sequence start enables deep, implicit interactions between the target and subsequent behaviors. This lightweight yet effective strategy achieves superior performance, underscoring the significance of target-aware sequential modeling in industrial settings.

\section{Conclusion} 

In this paper, we propose a novel user behavior model, named GTS-LUM, designed to enhance the prediction of future user behaviors in the telecommunications domain. The model employs a (multi-modal) encoder-adapter-LLM decoder framework, incorporating several telecom-specific enhancements. Specifically, we introduce a timestamp processing method to address challenges associated with diverse time granularities. The model supports multi-modal data inputs, including structured tables and behavior co-occurrence graphs, and aligns these with the information originated from a semantic tokenizer using a Q-former structure. Additionally, we implement a front-placed target-aware mechanism to emphasize the most relevant historical behaviors linked to the target label, thereby facilitating the capture of users' long-term behavioral patterns. Extensive experiments conducted on industrial datasets validate the feasibility and superiority of this end-to-end architecture, compared to previous LLM4Rec approaches. As part of future work, we aim to investigate more efficient tools to accelerate the inference process while maintaining the effectiveness of the results.

\clearpage
\bibliographystyle{plain}
\bibliography{reference}

\begin{thebibliography}{10}

\bibitem{ALBOUKAEY2020113779}
Nadia Alboukaey, Ammar Joukhadar, and Nada Ghneim.
\newblock Dynamic behavior based churn prediction in mobile telecom.
\newblock {\em Expert Systems with Applications}, 162:113779, 2020.

\bibitem{CALZADAINFANTE2020113553}
Laura Calzada-Infante, María Óskarsdóttir, and Bart Baesens.
\newblock Evaluation of customer behavior with temporal centrality metrics for churn prediction of prepaid contracts.
\newblock {\em Expert Systems with Applications}, 160:113553, 2020.

\bibitem{bgem32024}
Jianlv Chen, Shitao Xiao, Peitian Zhang, Kun Luo, Defu Lian, and Zheng Liu.
\newblock Bge m3-embedding: Multi-lingual, multi-functionality, multi-granularity text embeddings through self-knowledge distillation, 2024.

\bibitem{HLLM2024}
Junyi Chen, Lu~Chi, Bingyue Peng, and Zehuan Yuan.
\newblock Hllm: Enhancing sequential recommendations via hierarchical large language models for item and user modeling, 2024.

\bibitem{RecForest2022}
Chao Feng, Wuchao Li, Defu Lian, Zheng Liu, and Enhong Chen.
\newblock Recommender forest for efficient retrieval.
\newblock In S.~Koyejo, S.~Mohamed, A.~Agarwal, D.~Belgrave, K.~Cho, and A.~Oh, editors, {\em Advances in Neural Information Processing Systems}, volume~35, pages 38912--38924. Curran Associates, Inc., 2022.

\bibitem{P52022}
Shijie Geng, Shuchang Liu, Zuohui Fu, Yingqiang Ge, and Yongfeng Zhang.
\newblock Recommendation as language processing (rlp): A unified pretrain, personalized prompt \& predict paradigm (p5).
\newblock In {\em Proceedings of the 16th ACM Conference on Recommender Systems}, RecSys '22, page 299–315, New York, NY, USA, 2022. Association for Computing Machinery.

\bibitem{node2vec2016}
Aditya Grover and Jure Leskovec.
\newblock node2vec: Scalable feature learning for networks, 2016.

\bibitem{HowtoIndexItem2023}
Wenyue Hua, Shuyuan Xu, Yingqiang Ge, and Yongfeng Zhang.
\newblock How to index item ids for recommendation foundation models.
\newblock In {\em Proceedings of the Annual International ACM SIGIR Conference on Research and Development in Information Retrieval in the Asia Pacific Region}, SIGIR-AP '23, page 195–204, New York, NY, USA, 2023. Association for Computing Machinery.

\bibitem{jaintelecom2021}
Hemlata Jain, Ajay Khunteta, and Sumit Srivastava.
\newblock Telecom churn prediction and used techniques, datasets and performance measures: a review.
\newblock {\em Telecommunication Systems}, 76(4):613--630, April 2021.

\bibitem{LEARN2024}
Jian Jia, Yipei Wang, Yan Li, Honggang Chen, Xuehan Bai, Zhaocheng Liu, Jian Liang, Quan Chen, Han Li, Peng Jiang, and Kun Gai.
\newblock Learn: Knowledge adaptation from large language model to recommendation for practical industrial application, 2024.

\bibitem{kang2018selfattentivesequentialrecommendation}
Wang-Cheng Kang and Julian McAuley.
\newblock Self-attentive sequential recommendation, 2018.

\bibitem{li2024semanticconvergenceharmonizingrecommender}
Guanghan Li, Xun Zhang, Yufei Zhang, Yifan Yin, Guojun Yin, and Wei Lin.
\newblock Semantic convergence: Harmonizing recommender systems via two-stage alignment and behavioral semantic tokenization, 2024.

\bibitem{Blip22023}
Junnan Li, Dongxu Li, Silvio Savarese, and Steven Hoi.
\newblock {BLIP}-2: Bootstrapping language-image pre-training with frozen image encoders and large language models.
\newblock In Andreas Krause, Emma Brunskill, Kyunghyun Cho, Barbara Engelhardt, Sivan Sabato, and Jonathan Scarlett, editors, {\em Proceedings of the 40th International Conference on Machine Learning}, volume 202 of {\em Proceedings of Machine Learning Research}, pages 19730--19742. PMLR, 23--29 Jul 2023.

\bibitem{MMGRec2024}
Han Liu, Yinwei Wei, Xuemeng Song, Weili Guan, Yuan-Fang Li, and Liqiang Nie.
\newblock Mmgrec: Multimodal generative recommendation with transformer model, 2024.

\bibitem{STORE2024}
Qijiong Liu, Jieming Zhu, Lu~Fan, Zhou Zhao, and Xiao-Ming Wu.
\newblock Store: Streamlining semantic tokenization and generative recommendation with a single llm, 2024.

\bibitem{RecGPT4V2024}
Yuqing Liu, Yu~Wang, Lichao Sun, and Philip~S. Yu.
\newblock Rec-gpt4v: Multimodal recommendation with large vision-language models, 2024.

\bibitem{review2024}
Awais Manzoor, M.~Atif~Qureshi, Etain Kidney, and Luca Longo.
\newblock A review on machine learning methods for customer churn prediction and recommendations for business practitioners.
\newblock {\em IEEE Access}, 12:70434--70463, 2024.

\bibitem{Text2Tracks2024}
Enrico Palumbo, Gustavo Penha, Andreas Damianou, José Luis~Redondo García, Timothy~Christopher Heath, Alice Wang, Hugues Bouchard, and Mounia Lalmas.
\newblock Text2tracks: Generative track retrieval for prompt-based music recommendation, 2024.

\bibitem{Tiger2023}
Shashank Rajput, Nikhil Mehta, Anima Singh, Raghunandan Hulikal~Keshavan, Trung Vu, Lukasz Heldt, Lichan Hong, Yi~Tay, Vinh Tran, Jonah Samost, Maciej Kula, Ed~Chi, and Maheswaran Sathiamoorthy.
\newblock Recommender systems with generative retrieval.
\newblock In A.~Oh, T.~Naumann, A.~Globerson, K.~Saenko, M.~Hardt, and S.~Levine, editors, {\em Advances in Neural Information Processing Systems}, volume~36, pages 10299--10315. Curran Associates, Inc., 2023.

\bibitem{ChurnNet2024}
Somak Saha, Chamak Saha, Md.~Mahidul Haque, Md. Golam~Rabiul Alam, and Ashis Talukder.
\newblock Churnnet: Deep learning enhanced customer churn prediction in telecommunication industry.
\newblock {\em IEEE Access}, 12:4471--4484, 2024.

\bibitem{PPFCM2019}
E.~Sivasankar and J.~Vijaya.
\newblock Hybrid ppfcm-ann model: an efficient system for customer churn prediction through probabilistic possibilistic fuzzy clustering and artificial neural network.
\newblock {\em Neural Comput. Appl.}, 31(11):7181–7200, November 2019.

\bibitem{tablegpt22024}
Aofeng Su, Aowen Wang, Chao Ye, Chen Zhou, Ga~Zhang, Gang Chen, Guangcheng Zhu, Haobo Wang, Haokai Xu, Hao Chen, Haoze Li, Haoxuan Lan, Jiaming Tian, Jing Yuan, Junbo Zhao, Junlin Zhou, Kaizhe Shou, Liangyu Zha, Lin Long, Liyao Li, Pengzuo Wu, Qi~Zhang, Qingyi Huang, Saisai Yang, Tao Zhang, Wentao Ye, Wufang Zhu, Xiaomeng Hu, Xijun Gu, Xinjie Sun, Xiang Li, Yuhang Yang, and Zhiqing Xiao.
\newblock Tablegpt2: A large multimodal model with tabular data integration, 2024.

\bibitem{tang2018personalizedtopnsequentialrecommendation}
Jiaxi Tang and Ke~Wang.
\newblock Personalized top-n sequential recommendation via convolutional sequence embedding, 2018.

\bibitem{infonce2018}
Aaron van~den Oord, Yazhe Li, and Oriol Vinyals.
\newblock Representation learning with contrastive predictive coding, 2019.

\bibitem{LETTER2024}
Wenjie Wang, Honghui Bao, Xinyu Lin, Jizhi Zhang, Yongqi Li, Fuli Feng, See-Kiong Ng, and Tat-Seng Chua.
\newblock Learnable item tokenization for generative recommendation, 2024.

\bibitem{EAGER2024}
Ye~Wang, Jiahao Xun, Minjie Hong, Jieming Zhu, Tao Jin, Wang Lin, Haoyuan Li, Linjun Li, Yan Xia, Zhou Zhao, and Zhenhua Dong.
\newblock Eager: Two-stream generative recommender with behavior-semantic collaboration, 2024.

\bibitem{Colarec2024}
Yidan Wang, Zhaochun Ren, Weiwei Sun, Jiyuan Yang, Zhixiang Liang, Xin Chen, Ruobing Xie, Su~Yan, Xu~Zhang, Pengjie Ren, Zhumin Chen, and Xin Xin.
\newblock Content-based collaborative generation for recommender systems.
\newblock In {\em Proceedings of the 33rd ACM International Conference on Information and Knowledge Management}, CIKM ’24, page 2420–2430. ACM, October 2024.

\bibitem{ILM2024}
Li~Yang, Anushya Subbiah, Hardik Patel, Judith~Yue Li, Yanwei Song, Reza Mirghaderi, and Vikram Aggarwal.
\newblock Item-language model for conversational recommendation, 2024.

\bibitem{MLLMMSR2025}
Yuyang Ye, Zhi Zheng, Yishan Shen, Tianshu Wang, Hengruo Zhang, Peijun Zhu, Runlong Yu, Kai Zhang, and Hui Xiong.
\newblock Harnessing multimodal large language models for multimodal sequential recommendation, 2025.

\bibitem{HSTU2024}
Jiaqi Zhai, Lucy Liao, Xing Liu, Yueming Wang, Rui Li, Xuan Cao, Leon Gao, Zhaojie Gong, Fangda Gu, Michael He, Yinghai Lu, and Yu~Shi.
\newblock Actions speak louder than words: Trillion-parameter sequential transducers for generative recommendations, 2024.

\bibitem{NoteLLM2024}
Chao Zhang, Shiwei Wu, Haoxin Zhang, Tong Xu, Yan Gao, Yao Hu, Di~Wu, and Enhong Chen.
\newblock Notellm: A retrievable large language model for note recommendation, 2024.

\bibitem{NoteLLM22024}
Chao Zhang, Haoxin Zhang, Shiwei Wu, Di~Wu, Tong Xu, Xiangyu Zhao, Yan Gao, Yao Hu, and Enhong Chen.
\newblock Notellm-2: Multimodal large representation models for recommendation, 2025.

\bibitem{tinyllama2024}
Peiyuan Zhang, Guangtao Zeng, Tianduo Wang, and Wei Lu.
\newblock Tinyllama: An open-source small language model, 2024.

\bibitem{ijcai2019p600}
Tingting Zhang, Pengpeng Zhao, Yanchi Liu, Victor~S. Sheng, Jiajie Xu, Deqing Wang, Guanfeng Liu, and Xiaofang Zhou.
\newblock Feature-level deeper self-attention network for sequential recommendation.
\newblock In {\em Proceedings of the Twenty-Eighth International Joint Conference on Artificial Intelligence, {IJCAI-19}}, pages 4320--4326. International Joint Conferences on Artificial Intelligence Organization, 7 2019.

\bibitem{ECR2024}
Xiaoyu Zhang, Ruobing Xie, Yougang Lyu, Xin Xin, Pengjie Ren, Mingfei Liang, Bo~Zhang, Zhanhui Kang, Maarten de~Rijke, and Zhaochun Ren.
\newblock Towards empathetic conversational recommender systems.
\newblock In {\em 18th ACM Conference on Recommender Systems}, RecSys'24, pages 84--93. ACM, October 2024.

\bibitem{Zhang_2023}
Yuting Zhang, Yiqing Wu, Ran Le, Yongchun Zhu, Fuzhen Zhuang, Ruidong Han, Xiang Li, Wei Lin, Zhulin An, and Yongjun Xu.
\newblock Modeling dual period-varying preferences for takeaway recommendation.
\newblock In {\em Proceedings of the 29th ACM SIGKDD Conference on Knowledge Discovery and Data Mining}, KDD ’23, page 5628–5638. ACM, August 2023.

\bibitem{LLMKGRec2024}
Qian Zhao, Hao Qian, Ziqi Liu, Gong-Duo Zhang, and Lihong Gu.
\newblock Breaking the barrier: Utilizing large language models for industrial recommendation systems through an inferential knowledge graph, 2024.

\bibitem{DIEN2018}
Guorui Zhou, Na~Mou, Ying Fan, Qi~Pi, Weijie Bian, Chang Zhou, Xiaoqiang Zhu, and Kun Gai.
\newblock Deep interest evolution network for click-through rate prediction, 2018.

\bibitem{DIN2018}
Guorui Zhou, Chengru Song, Xiaoqiang Zhu, Ying Fan, Han Zhu, Xiao Ma, Yanghui Yan, Junqi Jin, Han Li, and Kun Gai.
\newblock Deep interest network for click-through rate prediction, 2018.

\end{thebibliography}

\end{sloppypar}
\end{document}